\def\year{2022}\relax
\documentclass[letterpaper]{article} 
\usepackage{aaai22}  
\usepackage{times}  
\usepackage{helvet} 
\usepackage{courier}  
\usepackage[hyphens]{url}  
\usepackage{graphicx} 
\usepackage{natbib}  
\usepackage{caption} 

\usepackage[switch]{lineno}  %

\usepackage{amsmath}
\usepackage{amsthm}
\usepackage{amssymb}
\usepackage{bbm}
\usepackage{mathtools}
\usepackage{mathrsfs}
\usepackage{dsfont}
\mathtoolsset{showonlyrefs}

\usepackage{courier} 

\usepackage{lipsum} 

\usepackage[space]{grffile} 

\usepackage{theoremref}
\usepackage{graphicx}
\usepackage{wrapfig}
\usepackage{xcolor}
\usepackage{colortbl}
\definecolor{dark-red}{rgb}{0.4,0.15,0.15}
\definecolor{dark-blue}{rgb}{0,0,0.7}

\newcommand{\topic}[1]{\textcolor{violet}{}}

\usepackage{algpseudocode}
\usepackage[vlined,linesnumbered,ruled,algo2e]{algorithm2e}
\SetKwProg{Fn}{Function}{}{}
\let\oldnl\nl
\newcommand{\nonl}{\renewcommand{\nl}{\let\nl\oldnl}}

\usepackage{multicol}
\usepackage{enumitem}
\usepackage[utf8]{inputenc} 
\usepackage[T1]{fontenc}    
\usepackage{url}            
\usepackage{booktabs}       
\usepackage{nicefrac}       

\title{ Neural Dependency Coding inspired Multimodal Fusion}
\author {
        
        Shiv Shankar
}
\affiliations {
    \{sshankar\}@cics.umass.edu
}

\begin{document}

\maketitle

\begin{abstract}
Information integration from different modalities is an active area of research. Human beings and, in general, biological neural systems are quite adept at using a multitude of signals from different sensory perceptive fields to interact with the environment and each other.
Recent work in deep fusion models via neural networks has led to substantial improvements over unimodal approaches in areas like speech recognition, emotion recognition and analysis, captioning and image description. However, such research has mostly focused on architectural changes allowing for fusion of different modalities while keeping the model complexity manageable.
Inspired by recent neuroscience ideas about multisensory integration and processing, we investigate the effect of introducing neural dependencies in the loss functions. Experiments on multimodal sentiment analysis tasks:  CMU-MOSI and CMU-MOSEI with different models show that our approach provides a consistent performance boost.
\end{abstract}

\section{Introduction}

Human beings perceive the world as a unified whole, not in individual sensory modalities. While traditionally different sensory models have been studied in isolation, it has been well recognized that perception operates via the integration of information from multiple sensory modalities. Research in multimodal fusion aims to achieve a similar goal in artificial models: integrate different unimodal representations into a unified common representation.

Heterogeneities across different modalities mean that learning multimodal representations face challenges like feature shifts, distributional effects, nuisance variation and other related challenges \citep{baltruvsaitis2018multimodal}. Effective fusion still is an open and challenging task. While in some areas like opinion analysis \citep{garcia2019multimodal, soleymani2017survey}, speech recognition, image processing \citep{xu2015multi, xu2013survey} and trait analysis \citep{park2014computational} fusion models have achieved substantial improvements over their unimodal counterparts ;in other areas like sentiment analysis \cite{rahman2020integrating} the improvements have been uninspiring.

Current research in deep multimodal fusion primarily deals with architectural improvements to create complex feature rich yet efficient representations \cite{zadeh2017tensor, liu2018efficient, hazarika2020misa}. Recently \cite{rahman2020integrating} used pre-trained transformer \citep{tsai2019multimodal, siriwardhana2020jointly} based models to achieve state-of the-art results on multimodal sentiment benchmark MOSI \cite{wollmer2013youtube} and MOSEI \cite{zadeh2018multimodal}.

Unlike these prior works, we do not focus not on architectural changes. Inspiring from work in multisensory neural processing, we utilize a concept known as \emph{synergy} or \emph{dependency} to train these models. Synergy is an information-theoretic concept related to mutual information (MI). The synergy between random variables $X$ and $Y$ refers to the unique mutual information that $X$ provides about $Y$. As it is based on conditional mutual information, synergy is able to capture relations between modalities that are difficult to capture via techniques like subspace alignment which implicitly assume linear dependencies.

\section{Preliminaries}
In this section, we give an overview of mutual information and existing work on deep multimodal fusion.

\subsection{Multimodal Fusion}

The problem in the most abstract terms is a supervised learning problem. We are provided with a dataset of $N$ observations $\mathcal{D} = { (x_i,y_i)_{i=1}^{N}} $. All $x_i$ come from a space $\mathcal{X}$ and $y_i$ from $\mathcal{Y}$. We are provided a loss function $L : \mathcal{Y} \times \mathcal{Y} \rightarrow \mathbb{R}$ which is the task loss. Our goal is to learn a model $\mathcal{F}_{\theta}: \mathcal{X} \rightarrow \mathcal{Y}$ such that the total loss  $\mathcal{L} =  \sum_i L(\mathcal{F}(x_i),y_i)$ is minimized.
In multimodal fusion the space of inputs $\mathcal{X}$ naturally decomposes into $K$ different modalities $\mathcal{X} = \prod_{j=1}^K \mathcal{X}_j$.  We use $X_j$ to represent random variables which form the individual modality specific components of the input random variable $X$.

The standard way to learn such a multimodal function is to decompose it into two components: a) an embedding component $E$ which fuses information into a high dimensional vector in $\mathbb{R}^d$ and b) a predictive component $P$ which maps vector from  $\mathbb{R}^d$ to $\mathcal{Y}$. Furthermore since the different modalities are often no directly compatible with each other (for eg text and image), $E$ itself is decomposed into a) modality specific readers $F_i \mathcal{X}_i \rightarrow \mathbb{R}^{d_i}$ which are specifically designed for each individual modality $\mathcal{X}_i$ and b) a fusion component $F : \prod_i \mathbb{R}^{d_i} \rightarrow \mathbb{R}^d$ which fuses information from eah individual modality embedding. $F$ is provided with uni-modal representations of the inputs $X_i = (X_1,X_2, \dots X_K)$ obtained through embedding networks $f_i$. $F$ has to retain both unimodal dependencies (i.e relations between features that span only one modality) and multi-modal dependency (i.e relationships between features across multiple modalities).

This decomposition has two advantages a) the individual modality reader can be pre-trained on the task at hand or even from a larger dataset (for example  BERT \citep{devlin2018bert} for language, Resnet \citep{he2016identity} for images ) which allows us to leverage wider modality specific information and b) often but not always each individual modality is in principle enough to correctly predict the output

Plethora of neural architectures have been proposed to learn multimodal representations for sentiment classification. Models often rely on a fusion mechanism \citep{khan2012color}, tensor factorisation \citep{liu2018efficient, zadeh2019factorized} or complex attention mechanisms \citep{zadeh2018memory} that is fed with modality specific representations.

\subsection{Mutual Information}
Mutual information (MI) measures the dependence between two random variable $X$ and $Y$ and is able to incorporate all forms of relationships between the two. MI between $X$ and $Y$, is given by 
$$
I(X ; Y) = \mathbb{E}_{X Y}\left[\log \frac{p_{X Y}(x, y)}{p_{X}(x) p_{Y}(y)}\right]
$$
where $p_{X Y}$ is the joint probability density of the pair $(X, Y)$, and $p_{X}, p_{Y}$ are the marginal probability densities of $X,Y$ respectively \cite{cover1999elements}. It is also clear from the above expression that it equals the KL divergence of the joint density of $(X, Y)$  relative to the product of marginals of $X$ and $Y$.
\begin{align}
I(X ; Y) = KL\left[ p_{X Y}(x, y) \| p_{X}(x) p_{Y}(y)\right ] {\label{eqn:MIKL}}
\end{align}
Since under the assumption of independence, the distribution of the pair $(X,Y)$ is given by the product of their individual distributions; the MI between $X,Y$ is 0 if and only if $X,Y$ are independent. As such mutual information can be understood as a measure of the inherent dependence expressed in the joint distribution of $X,Y$. 

\paragraph{Estimation of Mutual Information}
The computation of MI and other mutual dependency measures based only on samples is difficult. “Reliably estimating mutual information from finite continuous data remains a significant and unresolved problem” \citep{kinney2014equitability}.
Recently several estimators have been proposed based on optimization of variational bounds with flexible neural networks as proposed witness functions. For example \citet{belghazi2018mine} maximize the Donsker-Varadhan bound \citep{donsker1985large} to estimate KL divergence. 

\subsection{Alternative Dependence Measures}
Given the interpretation of the mutual information between variables as a divergence between their joint distribution and the independence implied distribution, other measures of mutual dependence can and have been proposed by replacing the KL divergence in with other proper discrepancy measures \cite{lopez2018generalization,belghazi2018mine,griffith2014intersection}

We will focus on the Hilbert-Schmidt criteria (also knowns as MMD criteria) which is also used in our experiments
\begin{itemize}
    
    \item
    Hilbert-Schmidt Dependence \citep{gretton2005measuring} is obtained by using the Maximum Mean Discrepancy or MMD \citep{gretton2012mmd} instead of KL divergence in \eqref{eqn:MIKL}. Further extensions to MMD have been developed based on neural networks which provide non-universal but more powerful kernel based tests \citep{liu2020learning}. 
    $$I_{\mathcal{MMD}} = \mathcal{MMD}\left(p_{X Y}(x, y) || p_{X}(x) p_{Y}(y)\right)$$
    
\end{itemize}

\subsection{Neural Synergy}
The aforementioned measures have been developed for the case of two random variables. Extension of mutual information to the multivariate case is an active area of research in information theory \cite{griffith2014quantifying, ince2017measuring}. 
There is a vast literature on partial information decomposition (PID) \cite{williams2010nonnegative} which aims to characterize and analyze different divisions of mutual information into a set of non-negative terms. Several axiomatic frameworks have been proposed and analysed  to analyze the concepts of redundancy \citet{williams2010nonnegative, kolchinsky2019novel} and surprise \cite{ince2017measuring}. Variants of MI based on non-Shannon divergences \cite{james2017multivariate,gacs1973common} are less common but have also been analyzed in the PID framework. We will focus on a specific measure known as dual correlation TSE measure \cite{tononi1994measure}. This has been used by \citet{rosas2019quantifying,barrett2011practical} to analyze neural complexity and brain functional connectivity. It is sometimes also known by the name of synergy in the neuroscience literature \citep{mediano2019measuring} and we shall use TSE and synergy interchangeably


\section{Neural Dependency Coding in Multisensory Processing}
A common and vital feature of nervous systems is the integration of information arriving simultaneously from multiple sensory pathways.
The underlying neural structures have been found to be related in both vertebrates and invertebrates. The classic understanding of this process is that different sensory modalities are processed individually and then combined in various multimodal convergence zones, including cortical and subcortical regions \citep{ghazanfar2006neocortex}, as well as multimodal association areas \citep{rauschecker1995processing}. 
Studies in the superior colliculus \citep{meredith1987determinants} showed that multiple sensory modalities are processed in this brain stem region, with some neurons being exclusively unimodal and others being multimodal. Hypotheses of encoding of multimodal information include changes in neuronal firing rates \citep{pennartz2009identification} or a combinatorial code in population of neurons \citep{osborne2008neural,rohe2016distinct}.



Studies of multisensory collicular neurons suggest that their crossmodal receptive fields (RF) often overlap \cite{spence2004crossmodal}. This pattern is also found in multisensory neurons present in other brain regions. As such, a spatiotemporal hypothesis of multisensory integration has been suggested: superadditive multimodal processing is observed when information from different modalities comes from spatiotemporally overlapping receptive fields \citep{recanzone2003auditory,wallace2004visual, stanford2005evaluating}. Since multimodal cortical neurons are generally downstream of modality-specific regions, the information about RF overlap is present in their input unimodal neural representations. Recent observations have led to the discovery of multimodal neurons in the generally modality-specific regions suggesting that a non-trivial part of the process happens in distributed circuits \citep{schroeder2005multisensory,stein2008multisensory}. \citet{tyll2011thalamic} provide evidence that that cortical multimodal processing is influenced via corticothalamic connections. Moreover, the sensory-specific nuclei of the thalamus have been shown to feed multisensory information to primary sensory specific-cortices \citep{kayser2008visual}

Other evidence also shows that while multimodal representations are distinct from unimodal ones, there is sufficient overlap between the set of neurons that process different sensory modalities. For example, \citet{follmann2018multimodal} show that even in a simple crustacean organism, more than half the neurons in the commissural ganglion are multimodal. Moreover, they show that in 30\% of these multimodal neurons, responses to one modality were predictive of responses to other modalities. Both these facts suggest that the neural representations across different modalities have high information about each other.

Cortical and subcortical networks often contain clusters of strongly connected neurons. Functionally the existence of such cliques imply highly integrated pyramidal cells that handle a disproportionately large amount of traffic \citep{harriger2012rich}. In cortical circuits, around 20\% of the neurons account for 80\% of the information propagation \citep{nigam2016rich, van2011rich}. \citet{timme2016high, faber2019computation} demonstrate that multimodal computation tends to concentrate in such local cortical clusters. They also found significantly elevated synergy in such clusters and that the amount of synergy was proportional to the amount of information flow, suggesting that neural synergy emerges 
where there is more significant cognitive processing and information flow. 

High synergy has also been hypothesized to facilitate inter-circuit communication and intra-circuit processing. Correlated neural representation, especially synergic activity, are indicative of the recurrent oscillations that are believed to underlie multisensory cognition\citep{ hasselmo2002proposed, hernandez2020medial, honey2017switching}. \citet{fries2015rhythms, yu2008small} show the importance of synergy for organizing information in cortical circuits. The synchronization of neuronal circuits is related to higher-order processing \citep{vinck2015arousal} especially in thalamic and cortical circuits. Moreover, \citet{sherrill2021synergistic} recently have shown that recurrent information flow in cortical circuits leads to an increase in neural synergy and neural complexity. 

\citet{sherrill2020correlated} show that correlated activity and neural synergy were positively related when multiple external correlated stimuli are provided. Thus, synergy in neural firings is a representation of the similarity between inputs. However, synergy is also low when the stimuli are very highly correlated. This suggests that correlated neural firings are a means of combining both redundancy and complementarity when faced with multiple inputs. This also points to the possibility that multimodal perception is most efficient when inputs produce maximal synergy.

\section{Model}

For our purposes we will limit ourselves to talk about tasks similar to the MOSI dataset. In this setting the input has three modalities viz audio ($a$), visal ($v$) and text ($t$) The fusion problem involves learning a representation $E_{\theta}$ that combined the uni-modal representations of the inputs $X_{a, v, t}=(X_{a}, X_{v}, X_{t})$.

To train our neural architecture we need to estimate the previously defined synergy measures. Multivarite synergy is defined in terms of mutual information between different random variables in the combined distribution. We extend synergy to also include an MMD based measure of dependence as well as defined earlier. For estimation of these dependence measures, we use estimators of the following distribution divergences for this purpose in our experiments. 

\begin{itemize}
    \item Kullback Liebler Dependence/Mutual Information -  We use the method of \citet{belghazi2018mine} who use the Donsker Varadhan bound \citep{donsker1985large,belghazi2018mine} to estimate the KL divergence between the requisite distribution. This version corresponds to the standard definition of synergy.
    \item Neural Hilbert Schmidt Dependence - We use neural kernel augmented MMD \citep{liu2020learning} to estimate the divergence between the requisite joint distributions to estimate synergy.
\end{itemize}
\section{Experiments}

\subsection{Datasets}
We empirically evaluate our methods on two commonly used datasets for multimodal training viz CMU-MOSI and CMU-MOSEI.

CMU-MOSI \citep{wollmer2013youtube} is sentiment prediction tasks on a set of short youtube video clips. CMU-MOSEI \citep{zadeh2018multi} is a similar dataset consisting of around 23k review videos taken from YouTube. The output in both cases is a sentiment score in $[-3,3]$. For each dataset, three modalities are available; audio, video, and language text.

\subsection{Models}
We run our experiments with the following architectures:
\begin{itemize}
    \item Tensor Fusion Network or TFN \cite{zadeh2017tensor} combined information via pooling of a high dimensional tensor representation of multimodal features. More specifically it does a multimodal Hadamard product of the aggregated features with RNN based language features.
    \item Memory Fusion Network or MFN \citep{zadeh2018memory} incorporate gated memory-units to store multiview representations. It then performs an attention augmented readout over the memory units to combine information into a single representation.
    \item MAGBERT \citep{rahman2020integrating} is a transformer based architecture that uses the Wang gate \citep{wang2019words}. The multimodal information is send to the multimodal gate to compute modified embeddings which are passed to a BERT \citep{devlin2018bert} based model.
\end{itemize}

\subsection{Evaluation}
For our experiments we evaluate and report both the Mean Absolute Error (MAE) and the correlation of model predictions with true labels. Existing works \citep{rahman2020integrating} also use the output of the regression model, to predict a positive or negative sentiment on the data, using it as binary classifier. Using the same approach we report the accuracy or our models. In the results Accuracy $Acc_{7}$ denotes accuracy on 7 classes and $Acc_{2}$ denotes the binary accuracy. We also report the correlation of model intensity predictions with true values.

\subsection{Results}

Results on MOSI are presented in Table \ref{tab:mosi} while Table \ref{tab:mosei} present results for MOSEI dataset.
We trained each of the models with the standard cross-entropy loss (labeled as MLE) and with cross-entropy loss regularized with the synergy penalty discussed earlier. On both datasets our model performs improve on their respective baselines; sometimes by more than 3 points.
Also, note that just changing the training loss provides a reasonable improvement over the state-of-the-art model MAGBERT \citep{rahman2020integrating}. This shows that our approach is generalizable across architectures.

Our results also show that in general MMD based models tends to be better than the KL divergence based models. The greater efficacy of MMD based synergy might be due to the inherent behavior of the MMD dependency, which is always well defined, or it might reflect the hardness of information estimation. For example, it is well known that reasonable bounds on standard mutual information are challenging to obtain \citep{kinney2014equitability}; while MMD estimators do not suffer from the entropy estimation issue and are consistent \citep{gretton2012mmd}

Recently \citet{colombo2021improving} has conducted experiments on similar lines. The main differences between the our method and their method are a) our method focuses on synergy terms whereas their proposal is optimizing joint mutual information between different unimodal representations; and b) they experiment with variational measures of information whereas we use exact measures in the MMD criteria. We replicate our experiments with their best performing model and present the results in our Tables \ref{tab:mosi, tab:mosei} with the label $\text{MI}_{Was}$. It is clear that our proposal is better than the Waserstein model used by \citet{colombo2021improving}. Such an approach was also used by \citet{han2021improving}; however their proposal includes architectural changes as well which makes the exact comparison unclear. However these results present further evidence of the utility of some of form of dependence driver optimization for multimodal fusion.

\begin{table}[htb]
\begin{tabular}{|l|llll|}
\hline
   & $Acc_{7}$   & $Acc_{2}$   & MAE  & CORR \\
\hline
\hline
   & & \multicolumn{2}{c}{MFN} &  \\
\hline
MLE & 31.3 & 76.6 & 1.01 & 0.62 \\
MLE+$\text{S}_{KL}$ & 34.5 & 76.9 & 0.94 & 0.65 \\
MLE+$\text{S}_{MMD}$ & 35.9 & 77.4 & 0.95 & 0.66 \\
$\text{MI}_{Was}$ & 35.1 & 77.1 & 0.97 & 0.63 \\
\hline
   & & \multicolumn{2}{c}{LFN} &  \\
\hline
MLE & 31.9 & 76.9 & 1.01 & 0.64 \\
MLE+$\text{S}_{KL}$ & 32.6 & 77.6 & 0.97 & 0.64 \\
MLE+$\text{S}_{MMD}$ & 35.4 & 77.9 & 0.97 & 0.67 \\
$\text{MI}_{Was}$ & 32.4 & 77.6 & 0.97 & 0.64 \\
\hline
   & &\multicolumn{2}{c}{MAGBERT} & \\
\hline
MLE & 40.2 & 83.7 & 0.79 & 0.80 \\
MLE+$\text{S}_{KL}$ & 41.9 & 84.3 & 0.76 & 0.82 \\
MLE+$\text{S}_{MMD}$ & 41.9 & 85.6 & 0.76 & 0.82 \\
$\text{MI}_{Was}$ & 41.8 & 84.2 & 0.76 & 0.82 \\
\hline
\end{tabular}
\caption{Results on sentiment analysis on CMU-MOSI. $Acc_{7}$ denotes accuracy on 7 classes and $Acc_{2}$ the binary accuracy. $MAE$ denotes the Mean Absolute Error and Corr is the Pearson correlation \label{tab:mosi}}
\end{table}

\begin{table}[htb]
\begin{tabular}{|l|llll|}
\hline
   &  $Acc_{7}$   & $Acc_{2}$   & MAE  & CORR \\
\hline
\hline
   & & \multicolumn{2}{c}{MFN} & \\
\hline
MLE &  44.3 & 74.7 & 0.72 & 0.52 \\
MLE+$\text{S}_{KL}$ &  45.3 & 74.8 & 0.72 & 0.56 \\
MLE+$\text{S}_{MMD}$ &  46.2 & 75.1 & 0.69 & 0.56 \\
$\text{MI}_{Was}$ &  45.1 & 75.2 & 0.72 & 0.54 \\
\hline
   & & \multicolumn{2}{c}{LFN} & \\
\hline
MLE &  45.2 & 74.3 & 0.70 & 0.54 \\
MLE+$\text{S}_{KL}$ &  46.1 & 75.3 & 0.69 & 0.56 \\
MLE+$\text{S}_{MMD}$ &  46.3 & 75.3 & 0.67 & 0.56 \\
$\text{MI}_{Was}$ &  45.9 & 75.1 & 0.69 & 0.55 \\
\hline
   & & \multicolumn{2}{c}{MAGBERT} & \\
\hline
MLE &  46.9 & 84.8 & 0.59 & 0.77 \\
MLE+$\text{S}_{KL}$ &  47.4 & 85.3 & 0.59 & 0.79 \\
MLE+$\text{S}_{MMD}$ &  47.9 & 85.4 & 0.59 & 0.79 \\
$\text{MI}_{Was}$ &  47.2 & 85.0 & 0.59 & 0.78 \\
\hline
\end{tabular}
\caption{Results on sentiment analysis on CMU-MOSEI. $Acc_{7}$ denotes accuracy on 7 classes and $Acc_{2}$ the binary accuracy. $MAE$ denotes the Mean Absolute Error and Corr is the Pearson correlation \label{tab:mosei}}
\end{table}

\section{Conclusions}
In this paper, we introduced the idea of synergy maximization. We experimented with different measures of synergy based on discrepancy measures such as KL divergene and MMD distance. We show that training with synergy can produce benefit on even state-of-the-art architectures. Similar experiments have been performed recently by \citet{han2021improving} and \citet{colombo2021improving} suggesting mutual information based approaches can be used to improve multimodal fusion.

\bibliography{mybib}



\end{document}